%% file: main.tex
\documentclass{bmvc2k}
\input{math_commands.tex}

\input{preamable}

\title{Prompt-Based Exemplar Super-Compression and Regeneration for Class-Incremental Learning}

\addauthor{Ruxiao Duan}{ruxiao.duan@yale.edu}{1}
\addauthor{Jieneng Chen}{jchen293@jhu.edu}{2}
\addauthor{Adam Kortylewski}{akortyle@mpi-inf.mpg.de}{3,4}
\addauthor{Alan Yuille}{ayuille1@jhu.edu}{2}
\addauthor{Yaoyao Liu}{lyy@illinois.edu}{5}

\addinstitution{
 Yale University
}
\addinstitution{
Johns Hopkins University
}
\addinstitution{
University of Freiburg
}
\addinstitution{
Max Planck Institute for Informatics
}
\addinstitution{
University of Illinois \\Urbana-Champaign
}

\runninghead{Duan et al.}{Prompt-Based Exemplar Super-Compression \& Regeneration}

\begin{document}
 \maketitle
 \input{sec/0}
 \input{sec/1}
 \input{sec/2}
 \input{sec/3}
 \input{sec/4}
 \input{sec/5}
 \bibliography{egbib}
\input{sec/appendix}
\end{document}

%% file: math_commands.tex
\usepackage{amsmath,amsfonts,bm}

\def\eqref#1{equation~\ref{#1}}

\def\1{\bm{1}}

\DeclareMathAlphabet{\mathsfit}{\encodingdefault}{\sfdefault}{m}{sl}
\SetMathAlphabet{\mathsfit}{bold}{\encodingdefault}{\sfdefault}{bx}{n}



%% file: preamable.tex
\usepackage{hyperref}
\usepackage{url}
\usepackage{amssymb}
\usepackage{multirow}
\usepackage{array}
\usepackage{cite}
\usepackage{comment}
\usepackage{arydshln}
\usepackage{xcolor}
\usepackage[linesnumbered,lined,vlined,ruled,commentsnumbered]{algorithm2e}
\usepackage{fontawesome}
\usepackage{graphicx}
\usepackage{booktabs}
\newcommand{\myparagraph}[1]{\noindent\textbf{#1}}
\usepackage[linesnumbered,ruled,vlined]{algorithm2e}
\usepackage{xcolor}

%% file: sec/0.tex
\begin{abstract}

Replay-based methods in class-incremental learning~(CIL) have attained remarkable success. Despite their effectiveness, the inherent memory restriction results in saving a limited number of exemplars with poor diversity. In this paper, we introduce PESCR, a novel approach that substantially increases the quantity and enhances the diversity of exemplars based on a pre-trained general-purpose diffusion model, without fine-tuning it on target datasets or storing it in the memory buffer. Images are compressed into visual and textual prompts, which are saved instead of the original images, decreasing memory consumption by a factor of $24$. In subsequent phases, diverse exemplars are regenerated by the diffusion model. We further propose partial compression and diffusion-based data augmentation to minimize the domain gap between generated exemplars and real images. PESCR significantly improves CIL performance across multiple benchmarks, e.g., $3.2\%$ above the previous state-of-the-art on ImageNet-100. The code is available at \url{https://github.com/KerryDRX/PESCR}.

\end{abstract}

%% file: sec/1.tex
\section{Introduction}
\label{sec:intro}

Ideally, AI systems should be adaptive to changing environments, where the data are continuously observed.
The AI models should be capable of learning concepts from new data while maintaining the ability to recognize the old ones.
In practice, AI systems often have constrained memory budgets, because of which most of the historical data must be abandoned.
However, deep AI models suffer from catastrophic forgetting when updated by abundant new data and limited historical data, as previous knowledge can be overridden by new information \citep{CF, CM}.
To study how to overcome catastrophic forgetting, the class-incremental learning~(CIL) protocol \citep{iCaRL} is established.
CIL assumes training samples from various classes are introduced to the model in phases, with previous data mostly discarded from memory.

CIL has enjoyed significant advancements \citep{EWC, SynapticIntelligence, Synapses, RiemannianWalk, MomentMatching, Lifelong, DER, FOSTER, MEMO, Dualnet}, among which replay-based methods \citep{iCaRL,LUCIR,AANets,RMM,DER,Mnemonics} stand out in terms of performance by employing a memory buffer to store a limited number of representative samples (i.e., exemplars) from former classes.
During subsequent learning phases, these exemplars are revisited to help retain previously learned knowledge.

Although replay-based methods demonstrate notable effectiveness, they are still restricted by two main drawbacks.
Firstly, since the exemplar set is much smaller compared to the new training data, the model is biased towards the new classes.
Secondly, the poor exemplar diversity leads to overfitting on old classes.
These two issues are essentially incurred by the lack of quantity and diversity of exemplars, respectively. Therefore, by tackling these two problems, all replay-based CIL methods can potentially be enhanced.

We consider these questions in CIL:
is it efficient to save old-class information as RGB images?
Can we compress the images into something more compact so that more information can be stored with the same memory consumption?
In this paper, we propose a novel approach named 
\emph{\textbf{P}rompt-based \textbf{E}xemplar \textbf{S}uper-\textbf{C}ompression and \textbf{R}egeneration} (\nolinebreak{PESCR}).
Instead of directly storing the previous images, we compress them into visual and textual prompts, e.g., edge maps and textual descriptions, and save these prompts in memory.
In subsequent phases, diverse high-resolution exemplars are regenerated from the prompts by leveraging an off-the-shelf pre-trained diffusion model, e.g., ControlNet~\citep{ControlNet}.

\input{figs/teaser}

Compared to traditional direct replay methods, PESCR enjoys increased \textbf{quantity} and enhanced \textbf{diversity} of the exemplar set (Figure~\ref{fig:teaser}).
Firstly, the exemplar quantity is boosted by compression: since the memory consumption of a $1$-bit edge map is merely $\frac{1}{24}$ that of its $8$-bit, $3$-channel RGB image counterpart, $24$ times more exemplars can be saved within the same memory budget.
We call this process super-compression, as this compression ratio is far beyond that of existing image compression algorithms, without even compromising the image quality.
Secondly, instead of relying on the images only from the dataset itself, we apply diffusion-based image generation to produce unseen samples with great diversity. This can be achieved simply by changing the random seed of the diffusion model.

However, utilizing generated images for CIL model training leads to a potentially huge domain gap between old exemplars and new data.
We propose two techniques, partial compression and diffusion-based data augmentation, to mitigate this problem, enabling the CIL model to properly benefit from the synthetic exemplars without the need to fine-tune the diffusion model on the target dataset.
Since the same pre-trained diffusion model can be directly downloaded from the public cloud at any time when necessary, we do not need to store the fine-tuned generator using our own memory.

Extensive experiments show that our PESCR achieves top performance on both fine-grained and coarse-grained classification datasets: Caltech-256 \citep{Caltech}, Food-101 \citep{Food}, Places-100 \citep{Places}, and ImageNet-100 \citep{deng2009imagenet}.
We fully investigate the effect of PESCR under different CIL settings and demonstrate that our approach achieves tremendous improvements compared to the state-of-the-art (SOTA) CIL method, e.g., substantially increasing the average learning from half accuracy on $11$-phase ImageNet-100 by $3.2\%$.

Our contributions can be summarized as follows.
We challenge the traditional manner of saving old class exemplars as RGB images in CIL and propose a memory-efficient data storage approach based on prompts, significantly increasing exemplar quantity with the same memory cost.
A general-purpose ControlNet is employed to regenerate diverse high-resolution images from prompts during incremental stages, without fine-tuning it on our target datasets.
We devise two techniques, partial compression and diffusion-based data augmentation, to alleviate the domain gap between generated exemplars and real images.
Extensive experiments are conducted on four classification datasets, two CIL protocols, seven CIL methods, and three budget settings to evaluate the performance of our approach.

%% file: figs/teaser.tex
\begin{figure}[t]
  \centering
   \includegraphics[width=\linewidth]{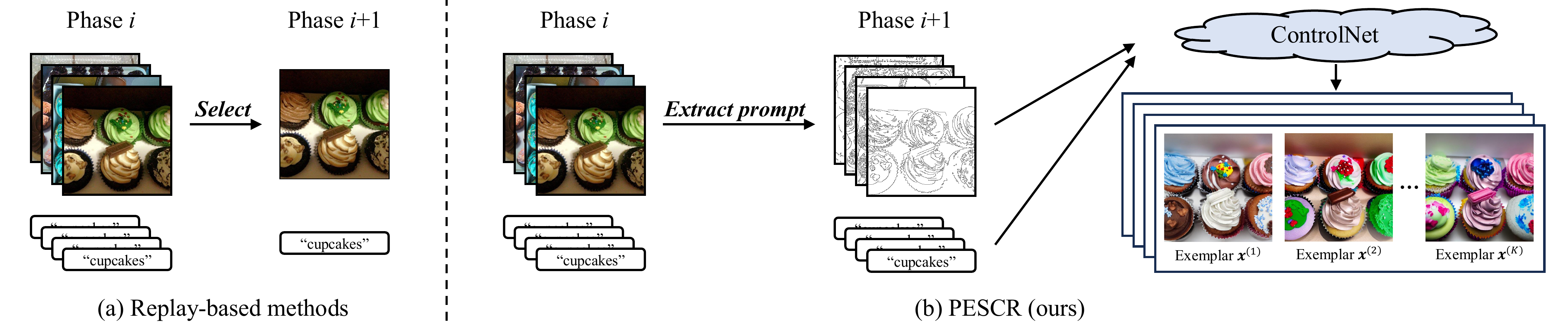}
   \caption{
   (a) \textbf{Traditional replay-based CIL methods} select and save only a small number of exemplars due to the memory restriction, leading to two severe issues:
   firstly, the relatively small size of the exemplar set compared to the new training dataset gives rise to a pronounced imbalance between old and new classes;
   secondly, the limited diversity of the exemplar set compared to the original training set incurs an overfitting problem.
   (b)~\textbf{PESCR} compresses old images into visual and textual prompts (edge maps and class tags) and saves prompts to the memory.
   In subsequent phases, diverse high-resolution exemplars are regenerated from these prompts via an off-the-shelf pre-trained diffusion model, e.g., ControlNet.
   PESCR improves the quantity and diversity of exemplars without violating the memory constraint.
   }
   \label{fig:teaser}
\end{figure}

%% file: sec/2.tex
\section{Related Work}
\label{sec:formatting}

\myparagraph{Class-incremental learning.}
The goal of continual learning~\citep{de2019continualsurvey,lopez2017gradient} or lifelong learning~\citep{chen2018lifelong,aljundi2017expert} is to develop machine learning models that can effectively adapt to and learn from data presented in a series of training stages.
Recent approaches on these topics are either task-based, where each new phase introduces data from all classes but belonging to a different task~\citep{DeepGenerative,WA,LwF}, or class-based, where each phase provides data from a new set of classes of the same dataset~\citep{Lifelong, DER, FOSTER, MEMO, Dualnet}.
The latter is typically called class-incremental learning (CIL).
CIL methods can be grouped into three main categories: replay-based, regularization-based, and parameter-based~\citep{de2019literaturereview,prabhu12356gdumb}. 
\textbf{Replay-based} methods employ a memory buffer to store information from previous classes for rehearsal in later stages.
Direct replay~\citep{Selective, Gradient, FeatureAdaptation, Rainbow, RMM, Placebos, OnlineHyperparameter} saves representative images from the dataset, while generative replay~\citep{DeepGenerative, ExemplarSupported, ModelAdaptation, Prototype, Fetril, LiuSHLSC21} saves generators trained by previous images.
These strategies are centered around selecting key samples, training generators, and effectively enhancing the classification model by utilizing a dataset that combines both exemplars and new data.
\textbf{Regularization-based} methods~\citep{EWC, SynapticIntelligence, Synapses, RiemannianWalk, MomentMatching} incorporate regularization terms into the loss function to reinforce previously acquired knowledge while assimilating new data.
\textbf{Parameter-based} methods~\citep{Lifelong, DER, FOSTER, MEMO, Dualnet} allocate distinct model parameters to each incremental phase, aiming to avoid model forgetting that arises from overwriting parameters.

\myparagraph{Exemplar compression.}
Attempts have been made to compress exemplars and reduce their memory consumption. MRDC \citep{MRDC} employs JPEG \citep{JPEG} to compress exemplars. CIM \citep{CIM} identifies foreground of exemplars by CAM \citep{CAM} and downsamples the background to compress images. These approaches have two weaknesses. 1) The compression is lossy, and the exemplar image quality degrades. 2) The compression ratio is data-dependent, as JPEG compression falls short with irregular patterns and non-smooth gradients, and background downsampling is inefficient with dominating foreground regions. Our PESCR, in contrast, guarantees high-resolution exemplars and a constant compression ratio of $24$, independent of the image dataset selected.

\myparagraph{Diffusion models with spatial control.}
Diffusion models function by progressively deteriorating the data, introducing Gaussian noise in incremental steps, and then learning to reconstruct the data by reversing the noise introduction process~\citep{Ho2020DiffusionModels,Singer2022Make,Villegas2022Phenaki}.
ControlNet~\citep{ControlNet} adds spatial conditioning controls to pre-trained text-to-image diffusion models, generating high-quality images from the input text and a visual prompt, which can be edges, depth, segmentation, or human pose representations.
This extraordinary ability sheds light on a new potential data format for storing old-class exemplars.
Since visual prompts such as edge maps cost much less space than the original RGB images, more prompts can be saved with the same memory budget, and more high-resolution images with the same details can be regenerated by ControlNet in later stages for CIL model training.

%% file: sec/3.tex
\section{Methodology}
\label{sec:method}

As illustrated in Figure~\ref{fig:framework}, our approach generates diverse exemplars from former prompts to overcome the forgetting problem in CIL.
We describe the problem setup of replay-based CIL in Section~\ref{subsec_setup}.
Then, we explain how to compress images to prompts in Section~\ref{subsec_exemplars}.
Next, we show how to regenerate the exemplars from prompts and train the CIL model with them in Section~\ref{subsec_training}.
Two techniques to reduce the domain gap are introduced in Section~\ref{sec:domain_gap}.

\subsection{Problem Setup}
\label{subsec_setup}

Reply-based CIL has multiple phases during which the number of classes gradually increases to the maximum~\citep{PODNet,LUCIR,Mnemonics}. In the {$1$-st} phase, we observe data $\mathcal{D}_{1}$, using them to learn an initial model $\Theta_1$. After training, we can only store a small subset of $\mathcal{D}_{1}$ (i.e., exemplars denoted as $\mathcal{E}_{1}$) in memory used as replay samples in later phases. In the $i$-th phase ({$i \geq 2$}), we get new class data $\mathcal{D}_{i}$ and load exemplars $\mathcal E_{1:i-1} = \mathcal E_1\cup \dots \cup \mathcal E_{i-1}$ from the memory. Then, we initialize $\Theta_i$ with $\Theta_{i-1}$, and train it using $\mathcal E_{1:i-1}\cup\mathcal{D}_{i}$. We evaluate the model $\Theta_{i}$ on a test set $\mathcal{Q}_{1:i}$ with all classes observed so far. Eventually, we select exemplars $\mathcal E_{1:i}$ from $\mathcal E_{1:i-1}\cup\mathcal{D}_{i}$ and save them in the memory. For the growing memory buffer setup with a budget of $b$ units per class (where one unit corresponds to the memory cost of one image), we need to ensure that the final number of exemplars stored at this stage $|\mathcal E_{1:i}| \leq ib$. For the fixed memory buffer setup with a budget of $B$ units in total, we need to have $|\mathcal E_{1:i}| \leq B$ in all phases.

\input{figs/framework}

\subsection{Prompt-Based Super-Compression}
\label{subsec_exemplars}

The performance of replay-based methods is severely limited by the quantity and diversity of exemplars.
To tackle these two issues, we compress old images into visual and textual prompts and use them to regenerate the exemplars by utilizing an off-the-shelf diffusion model.
As the prompts require much less memory compared to RGB images, we are able to save a large number of prompts within the same memory budget, so that far more exemplars can be regenerated in subsequent phases.

\myparagraph{Prompt extraction.}
At the end of the $i$-th phase, we first randomly select a training instance $(\boldsymbol{x}, y)$ from the dataset $\mathcal{D}_i$, where $\boldsymbol{x}$ denotes the image and $y$ denotes the classification label.
Next, we extract the visual prompts.
The visual prompts should preserve as many sufficient details as possible to help the CIL model retain the old-class knowledge.
They should also be small enough to reduce the memory cost.
We choose Canny edge maps as the visual prompts, using the classical Canny edge detector~\citep{CannyEdgeDetection} to obtain the edge map $\boldsymbol{e}$:
\begin{equation}
\label{eq_cannyedge}
    \boldsymbol{e} = \text{CannyEdge}(\boldsymbol{x}).
\end{equation}
Then, we directly save the class label $t$ as the textual prompt. For example, if the class label is ``cupcakes'', then $t=\text{``cupcakes''}$.
The above process is repeated to add other visual and textual prompts to the memory until the memory budget is exhausted.
After that, we obtain the final prompt set $\mathcal{P}_i$ for the $i$-th phase, i.e., $\mathcal{P}_i=\{(\boldsymbol{e}_j, t_j)\}_{j=1}^{R_i}$, where $R_i$ denotes the maximum number of prompts at the $i$-th phase to fit in the memory.

\myparagraph{Prompt memory consumption.} Storing edge maps and class tags in the memory buffer takes far less space than storing the original images.
For an $8$-bit RGB image, compressing it to a $1$-bit edge map of equal size achieves a compression ratio of $8 \times 3 = 24$. This means that each memory unit, which can originally store $1$ exemplar image, can now store $24$ edge maps. The class labels, which usually contain one or two words, consume negligible memory.

\subsection{Exemplar Regeneration and CIL Model Training}
\label{subsec_training}
To regenerate high-resolution exemplars from visual and textual prompts, we leverage a pre-trained ControlNet~\citep{ControlNet}, which is directly applicable without being fine-tuned on target datasets.
There is no need to allocate any memory space to save ControlNet, as we are able to re-download the ControlNet model from the cloud at the beginning of each phase.
This setting has its strengths and weaknesses.
By avoiding saving the generator in our buffer, we take full advantage of the memory to store old data.
However, we have to sacrifice the generator's capability to fit our dataset.
Consequently, the generated images might greatly differ from our images in various properties such as brightness, contrast, noise, etc., leading to a potentially huge domain gap.
We propose two solutions to tackle this issue in Section~\ref{sec:domain_gap}.

\myparagraph{Exemplar regeneration.}
In the $i$-th phase, we first take out a pair of visual and textual prompts $(\boldsymbol{e}, t)$ from the memory
$\mathcal{P}_{1:i-1} = \mathcal \mathcal{P}_1\cup \dots \cup \mathcal{P}_{i-1}$ of $R_{1:i-1} = \sum_{m=1}^{i-1} R_m$ prompts.
Then, we resize the edge map $\boldsymbol{e}$ with nearest neighbor interpolation to meet the input size requirement of ControlNet (both height and width must be a multiple of $64$). After that, we forward the prompts $(\boldsymbol{e}, t)$ to ControlNet and generate $K$ new exemplars with different random seeds:
\begin{equation}
\label{eq_controlnet}
   \hat{\boldsymbol{x}}_k = \text{ControlNet}(\boldsymbol{e}, t, s_k), \ \ k\in\{1,\cdots, K\},
\end{equation}
where $k$ and $s_k$ are the index and its corresponding random seed, respectively.
The generated image $\hat{\boldsymbol{x}}_k$ is then resized to the original size of $\boldsymbol{x}$ for consistency of image resolution.
We repeat this operation until we finish processing the entire prompt set $\mathcal{P}_{1:i-1}$.
Finally, we obtain the regenerated exemplar set $\hat{\mathcal{E}}_{1:i-1}$, which contains $R_{1:i-1} K$ exemplars.

\myparagraph{CIL model training.} Then, we combine the regenerated exemplars $\hat{\mathcal{E}}_{1:i-1}$ with the new training data $\mathcal{D}_{i}$ to train the CIL model:
\begin{equation}\label{eq_training_cil}
    \Theta_{i} \gets \Theta_{i} - \gamma \nabla_{\Theta_{i}}\mathcal{L}(\Theta_{i};\hat{\mathcal{E}}_{1:i-1}\cup\mathcal{D}_{i}),
\end{equation}
where $\gamma$ denotes the learning rate, and $\mathcal{L}$ is the CIL training loss.
To avoid bias towards a large number of similar regenerated images, in each epoch, we randomly sample only one image from the $K$ synthetic exemplars originating from the same base image for training.

\subsection{Exemplar-Image Domain Gap Reduction}
\label{sec:domain_gap}

With PESCR, we can significantly increase the quantity and diversity of exemplars without breaking the memory constraint.
However, the domain gap between the synthetic exemplars and real images is still a concern.
Therefore, directly using the regenerated exemplars for CIL model training leads to suboptimal performance in practice.
A common approach to solve this problem is to fine-tune the generative model on the target dataset.
However, if the generator is updated, we have to store it at the end of each phase, taking considerable buffer memory.
To circumvent this, we introduce the following two techniques to reduce the domain gap without fine-tuning ControlNet on our datasets.

\myparagraph{Partial compression.}
It might be tempting to utilize all memory budget to save edge maps, obtaining $24$ times the original number of exemplars.
However, excessive exemplars incur a greater domain gap.
Therefore, we consider only spending partial memory on saving edge maps, while leaving the rest to save RGB image exemplars as usual.
At each phase $i$, we assume the memory budget at this phase is $B_i$ units in total, i.e., we are allowed to save at most $B_i$ images as exemplars normally.
We set $\alpha$ as the compressed proportion of the dataset whose value can be adjusted depending on the CIL setting.
Hence, only $\alpha B_i$ memory units will be allocated for edge maps, saving $24 \alpha B_i$ edge maps; while the remaining $(1 - \alpha) B_i$ units will be allocated for images, saving $(1 - \alpha) B_i$ original images.

Herding \citep{iCaRL} is applied to sort the training images in $\mathcal{D}_i$ by their representativeness.
Then the $(1 - \alpha) B_i$ most representative ones are saved directly, while the next $24 \alpha B_i$ images are stored as edge maps in the buffer.
This preserves the most representative information of old classes in their original form.
The remaining less representative images in $\mathcal{D}_i$ are discarded.

\myparagraph{Diffusion-based data augmentation.}
During CIL model training, every real image $\boldsymbol{x}$ has a certain probability of being replaced by one of its $K$ generated copies.
Before training starts at the $i$-th phase, for each instance $(\boldsymbol{x}, y)$ with a real image $\boldsymbol{x}$, we extract the edge map $\boldsymbol{e}$ from $\boldsymbol{x}$ and obtain $K$ generated copies $\{\hat{\boldsymbol{x}}_k\}_{k=1}^K$ by Equation~\ref{eq_controlnet}.
In each training epoch, $\boldsymbol{x}$ has a probability $p$ of being replaced by any one of $\{\hat{\boldsymbol{x}}_k\}_{k=1}^K$ using uniform sampling.
This augmentation operation enables the model to learn from generated features and mitigates the domain gap between real and synthetic images.

\myparagraph{Quantity–quality trade-off.}
By adjusting the compressed proportion $\alpha$ and augmentation probability $p$, we can control the trade-off between the quantity and quality of generated information that is learned by the CIL model.
Larger $\alpha$ and $p$ let the model learn from more generated exemplars more frequently, but the widened domain gap might cause performance degradation.
Smaller $\alpha$ and $p$ alleviate the domain gap and improve the information quality, but the exemplar quantity and learning frequency are compromised.
The optimal choice of $\alpha$ and $p$ depends on the CIL setting.

\myparagraph{Overall CIL training loss.}
The CIL training process of the $i$-th phase (Equation~\ref{eq_training_cil}) is adjusted to: 
\begin{equation}\label{eq_training_cil_final}
    \Theta_{i} \gets \Theta_{i} - \gamma \nabla_{\Theta_{i}}\mathcal{L}(
        \Theta_{i};
        (\hat{\mathcal{E}}_{1:i-1} \vee {\mathcal{E}_{1:i-1}}) \cup (\hat{\mathcal{D}}_{i} \vee \mathcal{D}_{i})
    ),
\end{equation}
where $\hat{\mathcal{E}}_{1:i-1}$ and ${\mathcal{E}_{1:i-1}}$ represent the regenerated subset and the real-image subset of previous exemplars, respectively. $\hat{\mathcal{D}}_{i}$ and ${\mathcal{D}_{i}}$ are the augmented version and the original version of the new dataset at phase $i$, respectively. $\vee$ denotes the logic OR operation.





%% file: figs/framework.tex
\begin{figure}[t]
  \centering
   \includegraphics[width=\linewidth]{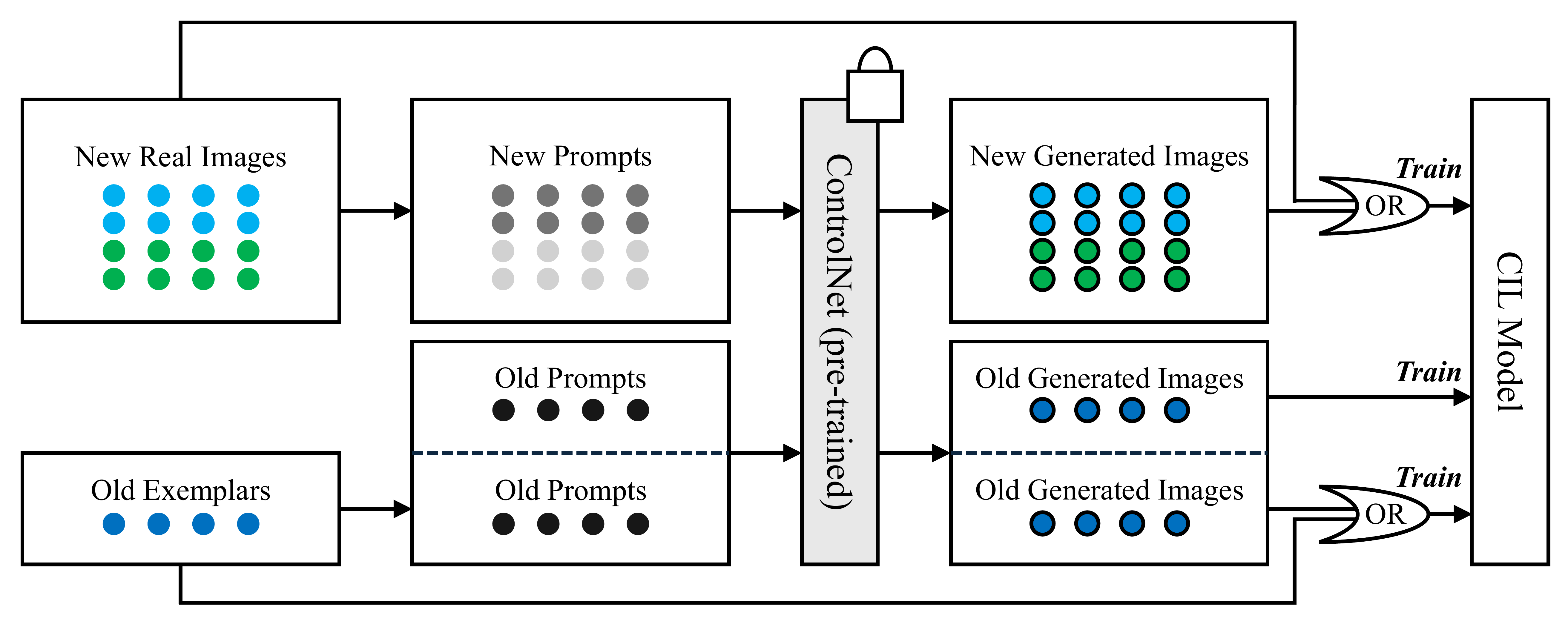}
   \caption{
   PESCR training process in the $i$-th phase ($i \geq 2$).
   A pre-trained ControlNet is downloaded in advance and remains frozen for image generation.
   Initially, we have three subsets of data: new real images of this phase, old prompts (i.e., edge maps and class tags), and old exemplars.
   Firstly, we transform the new real images and old exemplars into edge maps by Canny edge detection.
   Then, we use ControlNet to generate images from all the prompts we have.
   Finally, we train the CIL model with real and generated images.
   An image, if having both generated and real versions, will appear only once in each epoch, with a probability $p$ of being the generated version and $1-p$ of being the real version.
   }
   \label{fig:framework}
\end{figure}

%% file: sec/4.tex
\section{Experiments}

\subsection{Experiment Settings}
\label{subsec_settings}

\myparagraph{Datasets.} We conduct CIL experiments and evaluate PESCR on four image classification datasets: Caltech-256, Food-101, Places-100, and ImageNet-100.
\myparagraph{Caltech-256} \citep{Caltech} is an object recognition dataset with $30{,}607$ images from $257$ classes ($256$ object classes and a clutter class), each class having $80$ to $827$ images.
We remove the clutter class and keep at most $150$ images in each class by random selection to avoid extreme class imbalance.
Remaining images of each class are randomly split into training ($80\%$) and test ($20\%$) sets.
\myparagraph{Food-101} \citep{Food} contains $101{,}000$ food images of $101$ classes, each class with $750$ training and $250$ test images.
\myparagraph{Places-100} is a subset of Places-365-Standard \citep{Places}, a large-scale dataset including $365$ scene categories with $3{,}068$ to $5{,}000$ training and $100$ validation images per class.
We construct the subset by randomly choosing $100$ classes with seed $0$.
Then $3{,}000$ training images are randomly chosen from each category.
We use their validation set as the test set.
\myparagraph{ImageNet-100} is a subset of ImageNet-1000 \citep{deng2009imagenet} randomly sampled with seed $1993$, following \citep{LUCIR}, and each class has about $1{,}300$ training and $50$ test images.

\myparagraph{CIL protocols.}
We adopt two protocols in our experiments: learning from half (LFH) and learning from scratch (LFS).
LFH assumes the model is trained on half of the classes in the first phase and on the remaining classes evenly in the following $N$ phases.
LFS assumes the model learns from an equal number of classes in each of the $N$ phases.
We set $N$ to be $5$ or $10$ in our experiments.
The model is evaluated on all the classes observed so far at each phase, and the final average classification accuracy is reported.

\myparagraph{Memory budget.}
Although MEMO \citep{MEMO} recently proposes a new buffer setting (with model memory taken into account), based on which MEMO attains SOTA performance by storing lighter models, we still follow the common setup, assigning a fixed number of $b$ memory units per class for all methods.
The growing budget setting is more challenging than the fixed budget setting, where the memory in earlier phases is much more abundant.
(As Caltech-256 has fewer images per class, we set $b=5$ on Caltech-256 and $b=20$ on other datasets by default, unless otherwise specified.)

\myparagraph{Textual prompt extraction.}
Textual prompts are derived from class labels with minimal processing.
For Caltech-256, we remove the prefix and suffix and replace the hyphen with a space. e.g., ``063.electric-guitar-101'' is changed to ``electric guitar''.
For Food-101, we replace the underscore with a space. e.g., ``apple\_pie'' is modified as ``apple pie''.
For Places-100, which adopts a bi-level categorization scheme for some classes, such as ``general\_store/indoor'', ``general\_store/outdoor'', and
``train\_station/platform'', we transform them to ``indoor general store'', ``outdoor general store'', and ``train station platform'' to ensure semantic meaningfulness.
For ImageNet-100, the original class labels are used directly.

\myparagraph{Training setup.}
Algorithms are evaluated on ResNet-18 \citep{ResNet} which is trained by $200$ epochs in the first phase and $170$ epochs in subsequent ones with SGD.
Data augmentations include random resized cropping, horizontal flip, color jitter, and AutoAugment \citep{AutoAugment}, following \citep{FOSTER}.
Hyperparameters in PyCIL \citep{PyCIL} are adopted to implement all methods.
The previous SOTA exemplar compression approach CIM \citep{CIM} is also implemented by plugging into DER \citep{DER} and FOSTER \citep{FOSTER}, and we report the better result for each experiment.
Unless otherwise mentioned, we incorporate PESCR into DER \citep{DER}, which generally has the best performance across various settings.
We choose $\alpha$ and $p$ based on grid search and find that $\alpha \in [0.05, 0.3]$ and $p \in [0.2, 0.4]$ tend to work better in general.
$K = 5$ synthetic copies are generated per image for diffusion-based augmentation, but we do not train the model by more epochs.

\input{tables/lfs_and_lfh}
\input{tables/ablation}

\subsection{Results and Discussion}
\label{subsec_results}

\myparagraph{Comparison with previous approaches.}
We test the CIL performance of different methods with LFS and LFH protocols, illustrating the results in Table~\ref{tab:LFS_and_LFH}.
PESCR significantly enhances the baseline method DER by a large margin.
In $10$-phase LFS setting, PESCR improves accuracy by $5.0\%$ and $2.8\%$ on Caltech-256 and ImageNet-100, respectively.
In $11$-phase LFH setting, PESCR improves accuracy by $4.5\%$, $2.8\%$, $2.3\%$, and $3.2\%$ on Caltech-256, Food-101, Places-100, and ImageNet-100, respectively.

\myparagraph{Ablation study.}
We investigate the effect of partial compression (by setting $\alpha > 0$) and diffusion-based data augmentation (by setting $p > 0$) on PESCR in Table~\ref{tab:ablation}.
These two operations respectively focus on improving quantity and diversity of exemplars.
For straightforward representation, we directly show the number of real ($R$) and synthetic ($S$) exemplars per class, instead of $\alpha$.
The compressed ratio can be expressed as $\alpha = 1 - \frac{R}{b}$.
It can be observed that when augmentation is not applied, the improvement by increasing exemplar quantity is relatively limited. When augmentation is applied, as the domain gap is reduced, the model benefits much more from additional exemplars. Jointly applying these two techniques yields the best result. However, excessive exemplars (when $\alpha > 0.3$) harm the model performance, as the model is biased towards learning from the dominating synthetic data.

\myparagraph{PESCR with different CIL methods.} We incorporate PESCR with six different CIL methods and measure the accuracy increase in Table~\ref{tab:PESCR_with_different_baselines}.
PESCR improves the performance of five methods considerably by a large margin (from $3.2\%$ to $16.1\%$), and has relatively weak performance only on FOSTER~\citep{FOSTER}.
As reported in~\citep{FOSTER}, increasing the number of exemplars per class from $20$ to $50$ in FOSTER brings almost no improvement, but with more than $50$ exemplars, the domain gap leads to performance degradation for PESCR, thus the accuracy increase on FOSTER is not as tremendous as in other methods.

\myparagraph{PESCR with different memory budgets.}
To further understand the effectiveness of our approach under more limited memory limitations, we alter the memory budget $b$ and quantify the improvements of PESCR in Table~\ref{tab:budgets}.
It is remarkable that our approach can consistently enhance CIL performance even with more restricted budgets.

\input{tables/PESCR_with_different_baselines}
\input{tables/budgets}
\input{tables/Acc_vs_K}
\input{figs/gradcam}

\myparagraph{Number of generated copies per image.}
We explore the influence of increasing $K$ (the number of generated copies per image) on PESCR.
Table~\ref{tab:Acc_vs_K} illustrates the accuracy improvement by generating more copies, thanks to the growing sample diversity.
The performance increment gradually diminishes when $K \geq 5$, due to
1) the duplicated features generated from each edge map,
2) limited training epochs to learn from additional generations,
and 3) the widened domain gap between real and superfluous synthetic images.
Therefore, we use $K = 5$ copies per image for all the experiments.

\myparagraph{Grad-CAM visualization of generated exemplars.}
To verify if the class objects generated by ControlNet can be successfully identified by the CIL model, we use Grad-CAM \citep{GradCAM} to detect the image region with the most importance to the classification decision.
Displayed in Figure~\ref{fig:gradcam} are two synthetic exemplars of Food-101 with their activation maps produced by a model trained without diffusion-based data augmentation and a model trained with augmentation.
Evidently, the augmentation process is vital for the classification model to comprehensively detect the generated objects.
This ensures that the model can properly benefit from the synthetic exemplars during training in subsequent stages.

%% file: tables/lfs_and_lfh.tex
\begin{table}[t]
\caption{
Average $N$-phase LFS and $(N+1)$-phase LFH accuracies (\%) of different methods, with $b = 5$ memory units/class for Caltech-256 and $b = 20$ for other datasets.
}
\centering
\resizebox{\textwidth}{!}{
\begin{tabular}{@{\extracolsep{4pt}} c  *{4}{>{\centering\arraybackslash}m{9mm}} c *{8}{>{\centering\arraybackslash}m{9mm}}@{}}
\toprule[1pt]
    \multirow{3}{*}{CIL Method}
    &\multicolumn{4}{c}{Learning from Scratch (LFS)}
    &&\multicolumn{8}{c}{Learning from Half (LFH)}
\\
    \cline{2-5} \cline{7-14}
    &\multicolumn{2}{c}{Caltech-256}
    &\multicolumn{2}{c}{ImageNet-100}
    &&\multicolumn{2}{c}{Caltech-256}
    &\multicolumn{2}{c}{Food-101}
    &\multicolumn{2}{c}{Places-100}
    &\multicolumn{2}{c}{ImageNet-100}
\\
    \cline{2-3} \cline{4-5} \cline{7-8} \cline{9-10} \cline{11-12} \cline{13-14}
    {}
    &$N$=5
    &10
    &5
    &10
    &&5
    &10
    &5
    &10
    &5
    &10
    &5
    &10
\\  \hline
    iCaRL \citep{iCaRL}
    &57.7	&48.9	&66.2	&58.7
    &&53.4	&49.1	&58.6	&50.0	&42.2	&37.6	&59.2	&50.3
\\
    WA \citep{WA}
    &66.2	&54.6	&76.2	&69.7
    &&60.1	&46.8	&74.2	&64.6	&62.0	&57.3	&73.6	&66.2
\\
    MEMO \citep{MEMO}
    &65.7	&61.4	&78.5	&74.0
    &&62.6	&60.1	&71.1	&50.1	&53.4	&48.8	&72.5	&70.6
\\
    PODNet \citep{PODNet}
    &67.0	&60.9	&76.4	&68.7
    &&68.4	&66.6	&79.5	&77.0	&68.3	&66.3	&79.4	&77.2
\\
    FOSTER \citep{FOSTER}
    &41.3	&36.4	&81.1	&78.7
    &&62.4	&60.9	&81.2	&78.9	&69.4	&68.5	&81.6	&78.6
\\
    CIM \citep{CIM}
    &65.5	&66.3	&82.0	&78.0
    &&64.1	&65.5	&79.5	&77.1	&71.1	&70.5	&80.5	&79.5
\\
    DER \citep{DER}
    &68.1	&64.8	&81.8	&78.5
    &&68.4	&66.8	&82.0	&80.4	&70.4	&69.5	&81.8	&80.2
\\
    PESCR (ours)
    &\textbf{72.8}	&\textbf{69.8}	&\textbf{83.4}	&\textbf{81.3}
    &&\textbf{72.1}	&\textbf{71.3}	&\textbf{83.9}	&\textbf{83.2}	&\textbf{72.3}	&\textbf{71.8}	&\textbf{84.2}	&\textbf{83.4}
\\
\bottomrule[1pt]
\end{tabular}
}
\label{tab:LFS_and_LFH}
\end{table}

%% file: tables/ablation.tex
\begin{table}[t]
\caption{
Average $6$-phase LFH accuracies (\%) of PESCR on ImageNet-100, with $b=20$ memory units/class. Diffusion-based augmentation is applied with probability $p$. The numbers $R+S$ indicate $R$ real and $S$ synthetic exemplars/class are saved in the buffer.
}
\centering
\small
\begin{tabular}{c | *{9}{>{\centering\arraybackslash}m{9mm}}}
\toprule[1pt]
    $p$
    &20+0   &19+24  &18+48  &17+72  &16+96	&15+120  &14+144	&13+168	    &12+192
\\ \hline
    0.0
    &81.8
    &82.4
    &82.4
    &82.5
    &82.8
    &82.6
    &81.6
    &81.4
    &80.8
\\
    0.1
    &82.4
    &82.9
    &83.4
    &83.4
    &83.5
    &83.4
    &83.8
    &83.4
    &83.5
\\
    0.2
    &82.8
    &83.0
    &83.3
    &83.6
    &83.9
    &\textbf{84.2}
    &83.7
    &83.4
    &83.9
\\
    0.3
    &82.5
    &83.4
    &83.6
    &84.1
    &83.8
    &84.0
    &\textbf{84.2}
    &84.0
    &84.0
\\
    0.4
    &82.9
    &83.1
    &83.5
    &84.0
    &83.7
    &83.9
    &83.8
    &84.0
    &83.6
\\
    0.5
    &82.4
    &82.2
    &83.1
    &82.8
    &83.5
    &83.5
    &83.4
    &83.7
    &83.4
\\
\bottomrule[1pt]
\end{tabular}
\label{tab:ablation}
\end{table}

%% file: tables/PESCR_with_different_baselines.tex
\begin{table}[t!]
\caption{
Average $11$-phase LFH accuracies (\%) on ImageNet-100 with $b=20$ memory units/class, with and without PESCR plugged in.
}
\centering
\small
\begin{tabular}{c | cccccc}
\toprule[1pt]
    {}
    &iCaRL
    &WA
    &MEMO
    &PODNet
    &FOSTER
    &DER
\\ \hline
    Baseline
    &50.3
    &66.2
    &70.6
    &77.2
    &78.6
    &80.2
\\
    PESCR
    &66.4
    &71.7
    &77.6
    &80.6
    &79.4
    &83.4
\\ \hdashline
    Improvements
    &+16.1
    &+5.5
    &+7.0
    &+3.4
    &+0.8
    &+3.2
\\
\bottomrule[1pt]
\end{tabular}
\label{tab:PESCR_with_different_baselines}
\end{table}

%% file: tables/budgets.tex
\begin{table}[t!]
\caption{
Average $11$-phase LFH accuracies (\%) of DER on three datasets with $b$ memory units/class, with and without PESCR.
}
\centering
\resizebox{\textwidth}{!}{
\begin{tabular}{@{\extracolsep{4pt}} c  *{9}{>{\centering\arraybackslash}m{9mm}}@{}}
\toprule[1pt]
    \multirow{2}{*}{}
    &\multicolumn{3}{c}{Food-101}
    &\multicolumn{3}{c}{Places-100}
    &\multicolumn{3}{c}{ImageNet-100}
\\ \cline{2-4} \cline{5-7} \cline{8-10}
    {}
    &$b$=5
    &10
    &20
    &5
    &10
    &20
    &5
    &10
    &20
\\ \hline
    DER
    &77.3	&79.1	&80.4
    &67.1	&68.6	&69.5
    &77.9	&79.3	&80.2
\\
    DER+PESCR
    &80.1	&81.3	&83.2
    &70.4	&70.9	&71.8
    &81.1	&81.4	&83.4
\\ \hdashline
    Improvements
    &+2.8
    &+2.2
    &+2.8
    &+3.3
    &+2.3
    &+2.3
    &+3.2
    &+2.1
    &+3.2
\\
\bottomrule[1pt]
\end{tabular}
}
\label{tab:budgets}
\end{table}

%% file: tables/Acc_vs_K.tex
\begin{table}[t]
\caption{
Average $6$-phase LFH accuracies (\%) of PESCR on ImageNet-100 with $b=20$ memory units/class and $K$ generated copies per image. Diffusion augmentation is applied with probability $p=0.4$, and $R=16$ real and $S=96$ synthetic exemplars/class are saved.
}
\centering
\small
\begin{tabular}{c | *{7}{>{\centering\arraybackslash}m{10mm}}}
\toprule[1pt]
    $K$
    &0
    &1
    &5
    &10
    &15
    &20
    &25
\\ \hline
    Accuracy
    &81.8
    &83.1
    &83.7
    &83.4
    &\textbf{83.8}
    &83.7
    &83.4
\\
\bottomrule[1pt]
\end{tabular}
\label{tab:Acc_vs_K}
\end{table}

%% file: figs/gradcam.tex
\begin{figure}[t!]
\centering
\includegraphics[width=\linewidth]{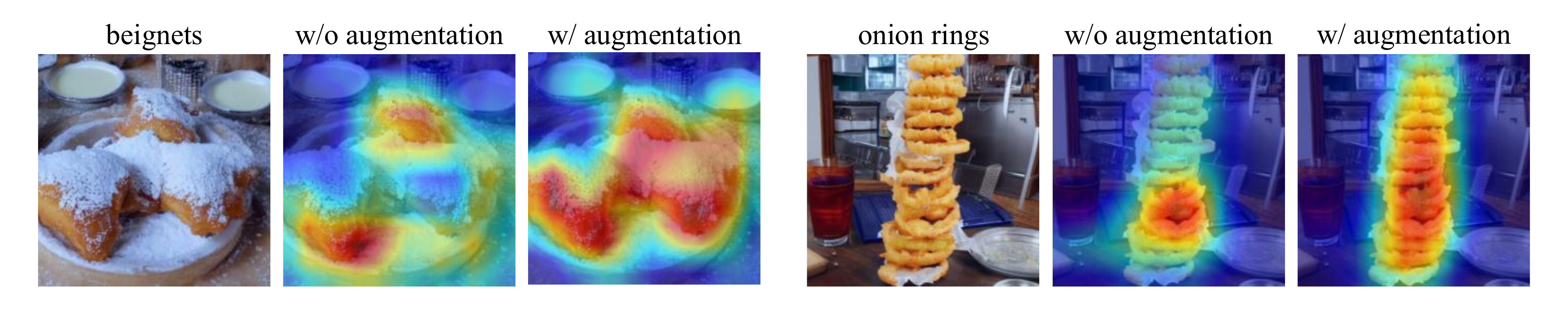}
\caption{
Two generated exemplars of Food-101 with Grad-CAM given by the classification model trained without diffusion-based augmentation and with augmentation.
Activation regions given by the model trained with augmentation capture class objects more accurately.
}
\label{fig:gradcam}
\end{figure}

%% file: sec/5.tex
\section{Conclusion}

We present PESCR, a novel prompt-based exemplar super-compression and regeneration approach that advances the state of replay-based class-incremental learning (CIL). By introducing a compact and expressive storage format based on visual and textual prompts, we challenge the conventional reliance on storing raw RGB images as exemplars. PESCR effectively compresses old-class information, dramatically increasing the number of stored exemplars within the same memory constraints. The prompts are later used to regenerate diverse, high-resolution synthetic images by an off-the-shelf pre-trained diffusion model, such as ControlNet, without requiring fine-tuning or additional memory storage for the generative model. To address the potential domain gap between generated and real images, we introduce two complementary strategies, partial compression and diffusion-based data augmentation, to allow PESCR to maintain model generalization and stability across phases.
Comprehensive experiments demonstrate that our approach constantly improves the model performance by a large margin on numerous datasets and CIL settings.

%% file: sec/appendix.tex
\clearpage
\setcounter{page}{1}
\setcounter{section}{0}
\setcounter{table}{0}
\setcounter{figure}{0}
\renewcommand\thetable{S\arabic{table}}
\renewcommand\thefigure{S\arabic{figure}}

\appendix

\section*{Supplementary Materials}

\myparagraph{Datasets.}
We present detailed information of the datasets used in our experiments in Table~\ref{tab:datasets}.

\myparagraph{Data preprocessing.}
The image transformation procedures are listed in Table~\ref{tab:transforms}.
Following FOSTER, we apply AutoAugment in all CIL methods for a fair comparison.
The same training transformations are applied to both real and generated images.

\myparagraph{Image resizing.}
There are two approaches to transform an image ($h \times w$) into an edge map ($H \times W$), which can be in a different size to accommodate the input requirement of ControlNet.
\emph{Resizing edge map}: convert the image ($h \times w$) to an edge map ($h \times w$) and then resize the edge map ($H \times W$) by nearest neighbor interpolation.
\emph{Resizing image}: resize the image ($h \times w$) into an intermediate image ($H \times W$) by Lanczos interpolation and then convert it to an edge map ($H \times W$).
We compare these two approaches qualitatively in Figure~\ref{fig:resize} and find that resizing the image can produce generations of higher quality.
These two approaches consume similar memory in total.

\myparagraph{Sampling scheme of prompt-based exemplars.}
We study the effect of sampling order for choosing exemplars to store as prompts.
The default sampling approach is to pick the most representative $(1-\alpha)b$ samples of each class to store as RGB images, while picking the next $24\alpha b$ representative ones to store as edge maps.
We refer to this scheme as \emph{least representative prompts}.
Its alternate sampling scheme, namely \emph{most representative prompts}, is to save the most representative $24\alpha b$ samples as prompts and preserve the next $(1-\alpha)b$ ones.
Another scheme is \emph{random sampling}, which picks $24\alpha b + (1-\alpha)b$ most representative samples first, and then randomly chooses $(1-\alpha)b$ to save as original images and $24\alpha b$ to save as prompts.
We compare these three sampling schemes in Table~\ref{tab:sampling_order}.
The three sampling modes yield similar accuracies, meaning that picking more or less representative images to be prompts does not have a significant impact on the final performance.

\myparagraph{Generation time.}
Image generation based on diffusion models is relatively time-consuming, which is a limitation of our approach.
For example, generating $K=1$ synthetic copy of ImageNet-100 using $8$ NVIDIA RTX A6000 GPUs takes approximately $15.5$ hours.
In this paper, we assume time is not a limiting factor in incremental learning, and we fully exploit the effectiveness of PESCR by increasing $K$.
In practical applications, if a time constraint is imposed, $K$ might be reduced for more efficient generation.

\myparagraph{Training time.}
We provide the time to train a CIL model (with DER) in Table~\ref{tab:time} with different numbers of exemplars per class.
By compressing $\alpha=10\%$ of the data, we can gain $66$ exemplars per class ($3.3$ times the original).
This costs approximately $32\%$ more training time, but the model accuracy is substantially increased from $81.9\%$ to $83.2\%$.

\myparagraph{Example generated images.}
From each dataset, we show two images with their class labels, Canny edge maps, and generations by ControlNet in Figure~\ref{fig:generations}.

\newpage
\input{tables/datasets}
\input{tables/transformations}
\input{figs/resize}
\input{tables/sampling_order}
\input{tables/time}
\input{figs/generations}

%% file: tables/datasets.tex
\begin{table}[ht!]
\caption{
Detailed dataset information including number of classes, total number of training/test images, average number of training/test images per class, and median image size.
}
\centering
\resizebox{\textwidth}{!}{
\begin{tabular}{c | cccc}
\toprule[1pt]
    Dataset
    &Caltech-256
    &Food-101
    &Places-100
    &ImageNet-100
\\ \hline
    \# classes
    &$256$
    &$101$
    &$100$
    &$100$
\\
    Total/average \# training images
    &$21{,}436$/$84$
    &$75{,}750$/$750$
    &$300{,}000$/$3{,}000$
    &$128{,}856$/$1{,}289$
\\
    Total/average \# test images
    &$5{,}472$/$21$
    &$25{,}250$/$250$
    &$10{,}000$/$100$
    &$5{,}000$/$50$
\\
    Median image size ($h\times w$)
    &$289\times300$
    &$512\times512$
    &$512\times683$
    &$375\times500$
\\
\bottomrule[1pt]
\end{tabular}
}
\label{tab:datasets}
\end{table}

%% file: tables/transformations.tex
\begin{table}[ht!]
\caption{
Training and test image transformations. Normalization has mean [0.485, 0.456, 0.406] and standard deviation [0.229, 0.224, 0.225].
}
\centering
\small
\begin{tabular}{@{\extracolsep{4pt}} c c @{}}
\toprule[1pt]
    Training transformations
    &Test transformations
\\ \hline
    RandomResizedCrop($224$),
    &Resize($256$), CenterCrop($224$),
\\
    RandomHorizontalFlip($p=0.5$),
\\
    ColorJitter(brightness=$63/255$),
\\
    ImageNetPolicy(),
\\
    ToTensor(),
    &ToTensor(),
\\
    Normalize(),
    &Normalize(),
\\
\bottomrule[1pt]
\end{tabular}
\label{tab:transforms}
\end{table}

%% file: figs/resize.tex
\begin{figure}[ht!]
  \centering
   \includegraphics[width=\linewidth]{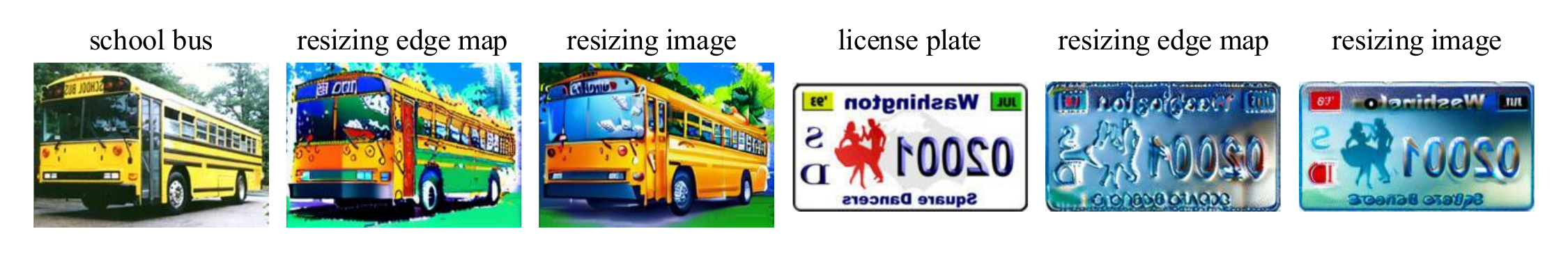}
   \caption{
    Two example images from Caltech-256 and their generated versions by resizing edge map and resizing image.
   }
   \label{fig:resize}
\end{figure}

%% file: tables/sampling_order.tex
\begin{table}[ht!]
\caption{
Average and last $6$-phase LFH accuracies (last accuracies in parentheses, \%) of ImageNet-100 with different exemplar sampling modes. Augmentation probability $p=0.4$, budget $b=20$ units/class, and $R=16$ real and $S=96$ synthetic exemplars/class are saved.
Each sampling mode is run three times and their average results are reported.
(LRP: least representative prompts; MRP: most representative prompts.)
}
\centering
\begin{tabular}{c | *{4}{>{\centering\arraybackslash}m{20mm}}}
\toprule[1pt]
    Sampling Mode
    &LRP
    &MRP
    &Random
\\ \hline
    Accuracy
    &\textbf{83.7} (79.7)
    &83.6 (79.6)
    &83.6 (79.7)
\\
\bottomrule[1pt]
\end{tabular}
\label{tab:sampling_order}
\end{table}

%% file: tables/time.tex
\begin{table}[ht!]
\caption{
Total training time ($2\times$A6000 GPU hours) and average LFH accuracies (\%) of DER on $11$-phase Food-101, with budget $b=20$ units/class and augmentation probability $p=0.4$.
$R$ real and $S$ synthetic exemplars/class are saved in the buffer.
}
\centering
\begin{tabular}{c | ccccc}
\toprule[1pt]
    \#Exemplars/Class=$R+S$
    &20=20+0   &43=19+24  &66=18+48  &89=17+72
\\ \hline
    Compressed Ratio $\alpha$
    &0.00
    &0.05
    &0.10
    &0.15
\\
    Training time
    &22.3
    &25.9
    &29.5
    &34.1
\\
    Accuracy
    &81.9	&82.0	&\textbf{83.2}	&82.4
\\
\bottomrule[1pt]
\end{tabular}
\label{tab:time}
\end{table}

%% file: figs/generations.tex
\begin{figure}[ht!]
  \centering
   \includegraphics[width=\linewidth]{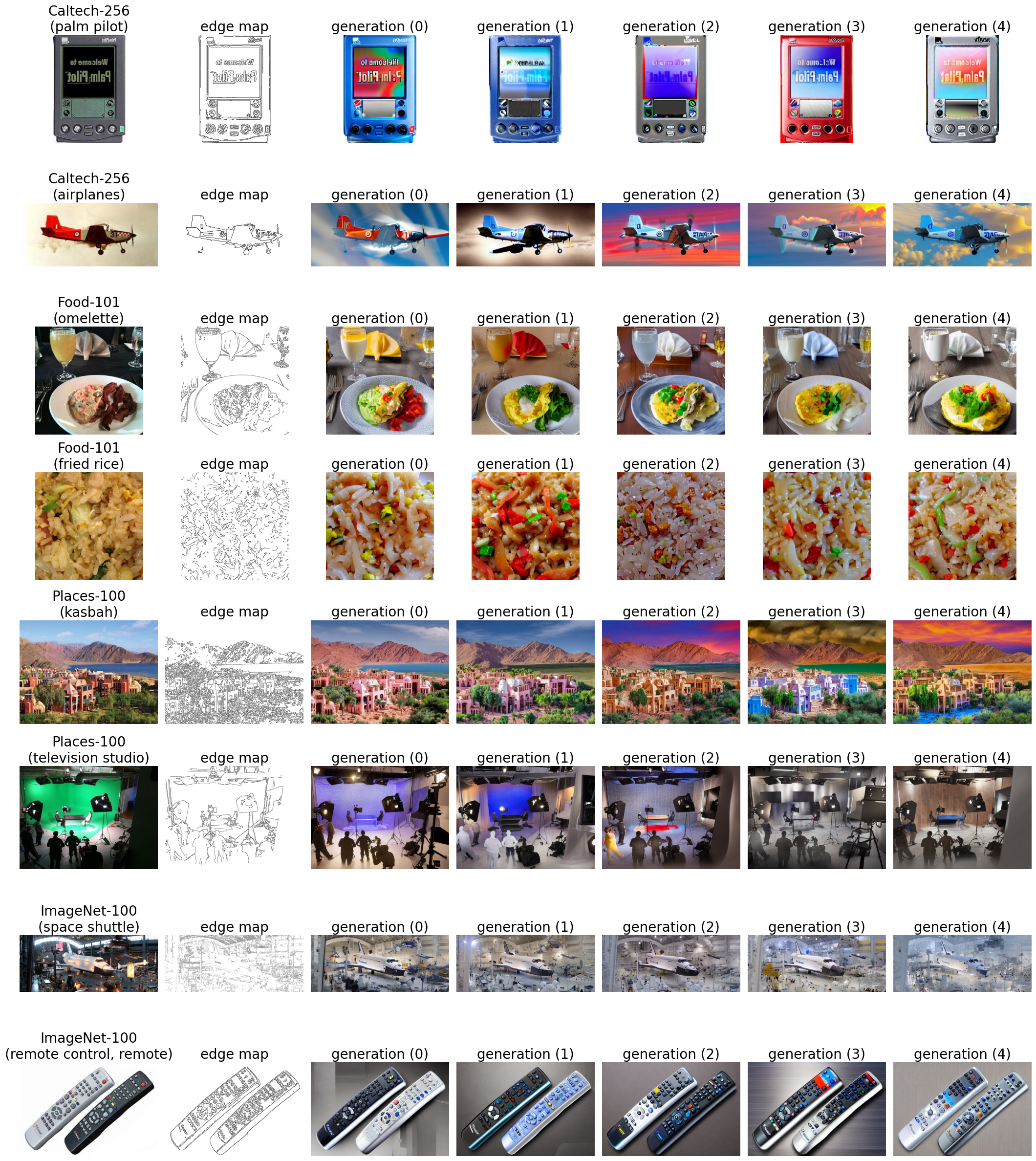}
   \caption{
    Random images selected from the four datasets, with their class labels, edge maps, and five generated images from ControlNet by random seeds $0,1,2,3,4$.
   }
   \label{fig:generations}
\end{figure}